\documentclass[fleqn,10pt]{wlscirep}
\usepackage[utf8]{inputenc}
\usepackage[T1]{fontenc}

\usepackage{amsmath}
\usepackage[table]{xcolor} 
\usepackage{subcaption}
\usepackage{graphicx}
\usepackage{svg}
\usepackage{booktabs}
\usepackage{xcolor}
\usepackage{url}
\usepackage{bm}

\title{HexagonalWarriorMamba: Superior Threshold-Dependent Multi-label Classification of 12-Lead ECG Cardiac Abnormalities}

\author[1]{Huawei Jiang}
\author[2]{Husna Mutahira}
\author[3]{Shibo Wei}
\author[4]{Jiahang Li}
\author[5]{Vladimir Shin}
\author[6]{Juneho Yi}
\author[3]{Dongryeol Ryu}
\author[3,*]{Wonyoung Park}
\author[7,*]{Mannan Saeed Muhammad}
\affil[1]{Sungkyunkwan University, Department of Computer Science and Engineering, Suwon, 16419, South Korea}
\affil[2]{Sogang University, Department of Computer Science and Engineering, Seoul, 04104, South Korea}
\affil[3]{Gwangju Institute of Science and Technology, Department of Biomedical Science and Engineering, Gwangju, 61005, South Korea}
\affil[4]{Tianjin Normal University, School of Artificial Intelligence, Tianjin, 300387, China}
\affil[5]{Financial University under the Government of the Russian Federation, Department of Artificial Intelligence, Moscow, 125167, Russia}
\affil[6]{Sungkyunkwan University, Department of Electrical and Computer Engineering, Suwon, 16419, South Korea}
\affil[7]{Queen Mary University of London, School of Electronic Engineering and Computer Science, London, E14NS, United Kingdom}
\affil[*]{corresponding author: parkwy0122@gist.ac.kr; mannan.muhammad@qmul.ac.uk}


\begin{abstract}
The accurate automated diagnosis of cardiac abnormalities from 12-lead electrocardiograms (ECGs) is critical for managing cardiovascular disease. However, detecting concurrent conditions remains a challenge for traditional deep learning models, which often have limited ability
to model the long-range dependencies inherent in ECG signals.
This manuscript proposes HexagonalWarriorMamba (HWMamba), a framework built on the Mamba architecture that processes 12-lead ECGs as single-channel 2D images rather than conventional 1D time series.
By integrating a hierarchical architecture with a 2D Selective Scan mechanism, HWMamba is designed to model global context and complex spatial relationships within the data.
The model is evaluated on the PhysioNet/Computing in Cardiology Challenge 2021 dataset, which includes 26 diagnostic labels and comprises recordings collected from seven institutions across four countries and three continents.
Results demonstrate that HWMamba outperforms current state-of-the-art (SOTA) methods across five key threshold-dependent metrics, including Challenge Score and Subset Accuracy.
These improvements provide a balance between strong discriminative capability and effective threshold selection derived from the training data, while maintaining near-SOTA performance in Macro AUROC. This Hexagonal Warrior performance, reflecting consistent performance across multiple evaluation dimensions, positions HWMamba as a robust and versatile approach for multi-label ECG classification.

\end{abstract}
\begin{document}

\flushbottom
\maketitle
%
%
\thispagestyle{empty}


\section*{Introduction}
Electrocardiography (ECG) measures the electrical activity of the heart from multiple perspectives, enabling non-invasive assessment of cardiac function. As cardiovascular diseases remain a leading global cause of mortality, the need for timely and accurate diagnostic tools is well established \cite{roth2017global}.
Furthermore, the increasing adoption of wearable devices and remote monitoring systems has led to the generation of large volumes of ECG data, necessitating efficient and scalable processing approaches \cite{prieto2022wearable}.

Large ECG datasets have made neural networks critical for detecting heart abnormalities \cite{ribeiro2020automatic, siontis2021artificial, abrar2025automated, dhandapani2025hybrid}. Automated interpretation reduces the workload on healthcare professionals and standardizes diagnosis. This is particularly valuable in resource-limited settings lacking expert cardiologists. 
%
Conventional models such as Convolutional Neural Networks (CNNs) and Transformers have demonstrated strong performance in this domain \cite{zhao2020ecg, che2021constrained}, but may face limitations when modeling the long sequences characteristic of ECG signals \cite{ansari2025survey}. More recently, selective State Space Models (SSMs), particularly Mamba \cite{gu2024mamba}, have been introduced as an alternative for handling long-range dependencies \cite{zhang2025switch}. Initially developed for language modeling, Mamba has since been extended to 2D data through VMamba \cite{liu2024vmamba}. VMamba employs a Cross Scan Module to model global context with improved computational efficiency compared to Transformer-based approaches.

This paper presents HexagonalWarriorMamba (HWMamba). In contrast to conventional approaches that rely on 1D time-series inputs \cite{nejedly2021classification, nejedly2022classification, jiang2025ecg, JiangECG}, the proposed method represents the 12-lead ECG as a single-channel image. This approach allows the use of established 2D image analysis architectures, including Swin Transformer \cite{liu2021swin}, ConvNeXt \cite{liu2022convnet}, and VMamba, to model both inter- and intra-lead relationships within the 12-lead configuration. Accordingly, HWMamba adopts a hierarchical structure and incorporates a 2D Selective Scan (SS2D) mechanism to efficiently process 2D feature maps.

\begin{figure}[t]
\centering
\includegraphics[width=0.5\textwidth]{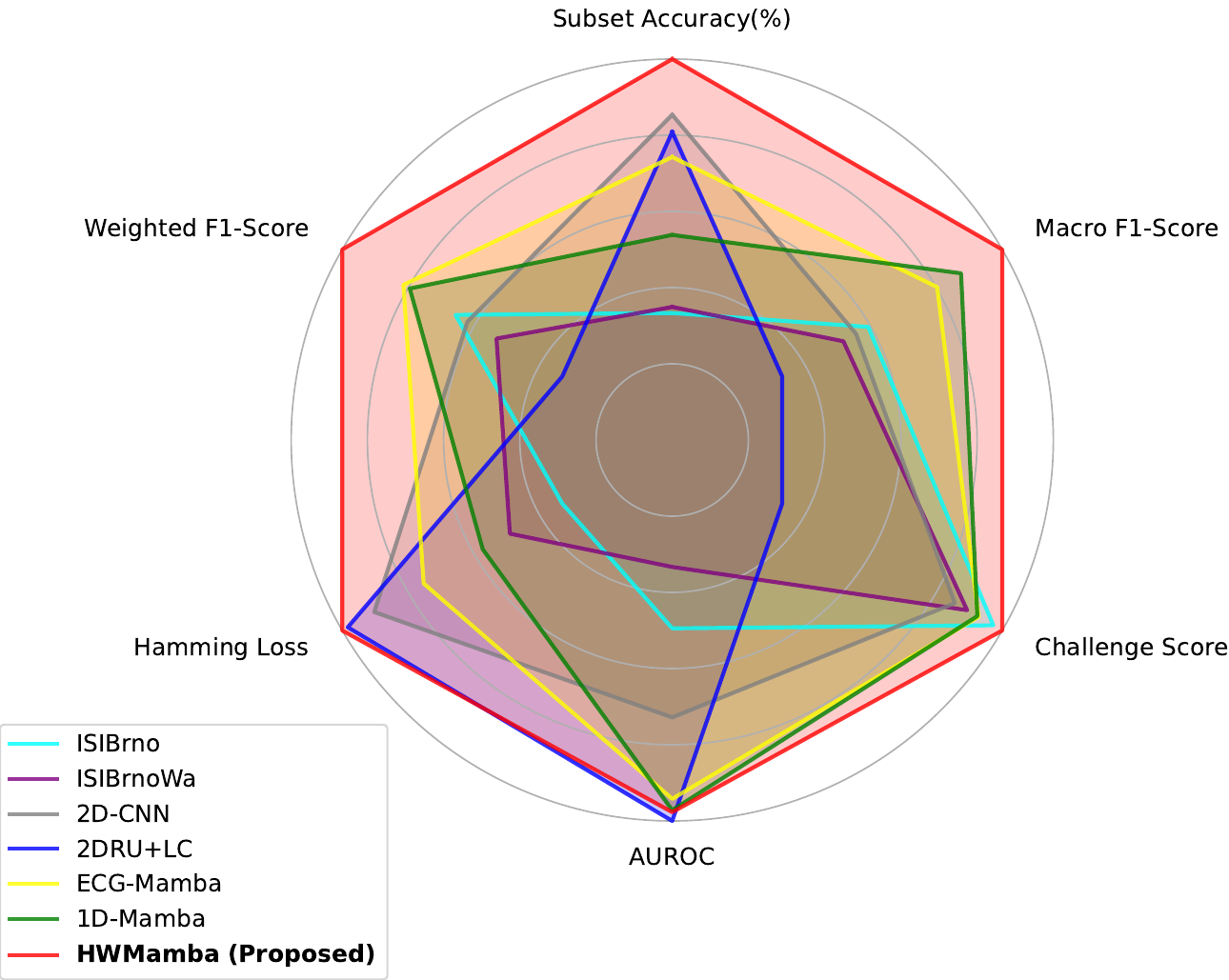}
\caption{Performance Comparison of Algorithms (Hexagon Radar Chart)} 
\label{fig1_hex}
\end{figure}

The proposed method was evaluated using publicly available datasets from the PhysioNet/Computing in Cardiology (CinC) Challenge 2021. The results indicate that HWMamba exhibits a balanced and consistent performance profile across multiple evaluation metrics. As illustrated in Figure \ref{fig1_hex}, the model achieves state-of-the-art (SOTA) performance on five threshold-dependent metrics, including Subset Accuracy, Challenge Score, Hamming Loss, Macro F1-score, and Weighted F1-score, while maintaining near-SOTA performance on the threshold-independent Macro AUROC metric.
The framework adopts a multi-label classification approach to better reflect clinical practice, where diagnoses often involve concurrent conditions \cite{li2025electrocardiogram, sun2024multitask}, for example, Atrial Fibrillation occurring alongside Right Bundle Branch Block \cite{zhang2022association, ran2023label}. In contrast to single-label classification, which selects only the highest confidence prediction, multi-label settings require threshold-based evaluation to assess multiple conditions independently. Within this context, HWMamba demonstrates a balance between strong discriminative capability and effective threshold selection derived from the training data. This behavior is associated with improved performance relative to recent methods such as 2DRU+LC \cite{hwang2024multi}.
The consistent performance observed across all six evaluation dimensions supports the suitability of HWMamba as a robust approach for multi-label ECG classification.


The contributions of this paper are summarized as follows: 
\begin{itemize}
    \item \textbf{Benchmarking and Performance:} This study establishes a new reference point for multi-label ECG classification, showing that HWMamba outperforms current SOTA methods on several threshold-dependent metrics, including Challenge Score and Subset Accuracy. This “Hexagonal Warrior” performance reflects a balanced and consistent performance profile across evaluation dimensions, supporting robust identification of concurrent cardiac abnormalities relative to existing approaches.
    \item \textbf{Generalization and Threshold Stability:} The analysis of the performance gap between training and testing thresholds indicates that strong performance on threshold-independent metrics such as AUROC does not necessarily translate to practical effectiveness. This highlights the importance of stable threshold selection for reliable multi-label cardiac diagnosis in real-world settings.
    
    
\end{itemize}

\section*{Related Work}
Recent advancements in the multi-label classification of 12-lead ECGs
can be broadly grouped into two dominant processing paradigms \cite{le2023scl}: treating the signal as a native 1D time series or transforming it into a structured 2D pseudo-image.

\subsection*{Classification on Native 1D Time-Series Input}

The most conventional approach involves processing 12-lead ECG data via 1D CNNs, utilizing the temporal nature of the electrical signals. A representative example of this methodology is the solution proposed by team ISIBrno \cite{nejedly2021classification}. Their method secured first place in the PhysioNet/CinC Challenge 2021, achieving the highest Challenge Score.
This method utilized an ensemble of residual networks enhanced with attention mechanisms to optimize feature extraction across leads.
Subsequent research by the same authors \cite{nejedly2022classification} revisited this architecture to examine the role of
attention modules. 
In contrast to the common trend of increasing model complexity, their findings suggested that a simplified residual network without attention achieved slightly improved performance compared to the original attention-based design \cite{nejedly2021classification}. This observation highlights the importance of careful architectural design when optimizing 1D backbones for cardiac signal analysis.
%

More recently, the focus has shifted toward SSMs to capture long-range dependencies in physiological signals. Both ECG-Mamba \cite{jiang2025ecg} and 1D-Mamba \cite{JiangECG} have adopted the Vision Mamba \cite{zhu2024vision} architecture adapted for 1D signal processing. The promising results demonstrated by these methods on the AUROC metric for multi-label heart disease classification validate the efficacy of architectures based on Mamba for analyzing ECG time series.

\subsection*{Classification via Structured 2D Pseudo-Image Input}

Parallel to 1D sequence modeling, a growing body of work explores the representation of 12-lead ECGs as rectangular 2D images to utilize established architectures \cite{liu2021swin, liu2022convnet} originally designed for computer vision.
Within this paradigm, the 2D-CNN approach \cite{elyamani2024deep} represents the 12-lead ECG as a 2D data structure with a single channel. This method utilizes Deep Residual 2D CNNs and demonstrates the effectiveness of spatial convolution operations in modeling correlations between leads.

Among recent approaches, the 2DRU+LC method \cite{hwang2024multi} reports strong performance on the threshold-independent AUROC metric. By adopting a U-Net architecture \cite{ronneberger2015u}, originally developed for biomedical image segmentation, this method introduces a structured design for ECG classification and achieves competitive results on AUROC.

While existing 2D approaches demonstrate strong performance on threshold-independent metrics, the proposed HWMamba framework is designed to address limitations in clinical applicability. In particular, it shows improved performance on threshold-dependent metrics, which are important for real-world diagnostic settings, while maintaining competitive performance on AUROC with only a modest difference relative to the 2DRU+LC method.

\section*{Methods}
\subsection*{Preliminaries}
This study utilizes the capability of SSMs for modeling long-range dependencies in ECG signals. Originating from classical control theory and the Kalman filter \cite{kalman1960new}, SSMs describe sequence modeling as a transformation of a 1D input signal $x(t) \in \mathbb{R}^{1 \times 1}$ to an output signal $y(t) \in \mathbb{R}^{1 \times 1}$ through a latent state representation $h(t) \in \mathbb{R}^{N \times 1}$.

\subsubsection*{State Space Models}
The fundamental continuous-time SSM is defined by two equations: a state equation \ref{state_eq} and an output equation \ref{output_eq}. Given an input signal $x(t)$, the system maps it to the latent state $h(t)$ and projects it to the output $y(t)$ as follows:
\begin{align}
    h'(t) &= \mathbf{A}h(t) + \mathbf{B}x(t) \label{state_eq} \\
    y(t) &= \mathbf{C}h(t) \label{output_eq}
\end{align}
Here, $\mathbf{A} \in \mathbb{R}^{N \times N}$ is the state matrix, while $\mathbf{B} \in \mathbb{R}^{N \times 1}$ and $\mathbf{C} \in \mathbb{R}^{1 \times N}$ serve as projection matrices. The $\mathbf{D}x(t)$ skip connection is excluded for simplicity.

\subsubsection*{Discretization}

Processing ECG feature maps requires discretizing the continuous-time state space model.
A timescale parameter $\Delta$ controls this transition by acting as the sampling step size. The Zero-Order Hold (ZOH) assumption treats the input as constant over each interval $\Delta$ and transforms the continuous parameters $(\mathbf{A}, \mathbf{B})$ into the discrete parameters $(\overline{\mathbf{A}}, \overline{\mathbf{B}})$ as follows:
\begin{align}
    \overline{\mathbf{A}} &= \exp(\bm{\Delta} \mathbf{A}) \\
    \overline{\mathbf{B}} &= (\bm{\Delta} \mathbf{A})^{-1}(\exp(\bm{\Delta} \mathbf{A}) - \mathbf{I}) \cdot \bm{\Delta} \mathbf{B}
\end{align}

By applying discretization, the SSM is adapted for discrete sequence data at each time step $k$ (see Supplementary Information for the complete mathematical proof).
\begin{align}
    h_k &= \overline{\mathbf{A}}h_{k-1} + \overline{\mathbf{B}}x_k \\
    y_k &= \mathbf{C}h_k
\end{align}

\subsubsection*{Selective State Space Models (Mamba)}
Mamba overcomes Linear Time-Invariant limitations through a selection mechanism computing the parameter set $\{\mathbf{B}, \mathbf{C}, \bm{\Delta}\}$ as functions of the input $x_k$ rather than remaining static:
\begin{align}
    \mathbf{B}_k &= \mathbf{W}_B x_k \\
    \mathbf{C}_k &= \mathbf{W}_C x_k \\
    \bm{\Delta}_k &= \text{softplus}(b_{\Delta} + \mathbf{W}_{\Delta} x_k)
\end{align}


where $\mathbf{W}_B, \mathbf{W}_C \in \mathbb{R}^{N \times D}$ and $\mathbf{W}_{\Delta} \in \mathbb{R}^{D \times D}$ are learnable projection matrices, with $D$ denoting the input channel dimension and $N$ the state dimension. The timescale projection $\mathbf{W}_{\Delta}$ is parameterized via a low-rank bottleneck of dimension $R$ (typically $R \ll D$) to improve computational efficiency. 
A softplus activation is then applied to enforce positivity, ensuring that the generated timescale $\bm{\Delta}$ represents a mathematically valid, nonnegative sampling interval. 
Additionally, $b_{\Delta}$ is a learnable bias parameter controlling the default timescale. Unlike $\bm{\Delta}$, the parameters $\mathbf{B}$ and $\mathbf{C}$ are generated via bias-free projections.


\subsection*{ECG Sample Pre-Processing}
All ECG signals were first uniformly resampled to a target rate of 500 Hz following established methodologies ~\cite{nejedly2021classification, nejedly2022classification, jiang2025ecg, JiangECG, natarajan2020wide}. Signals originally at 1000 Hz were downsampled using polyphase filtering to minimize aliasing. Conversely, 257 Hz signals were upsampled to 500 Hz using the Fast Fourier Transform. Signal length was standardized by right zero-padding shorter records to 8192 samples, while longer records were randomly sampled and truncated to this uniform length.

Normalization was intentionally omitted for the 12-lead ECG signals, following \cite{jiang2025ecg} and in contrast to several commonly adopted practices \cite{nejedly2021classification, nejedly2022classification, JiangECG, elyamani2024deep, hwang2024multi}. 
This decision preserves diagnostic information related to amplitude, which may otherwise be altered by standard normalization techniques such as Z-score scaling.
\begin{figure}[t]
\centering
\includegraphics[width=\textwidth]{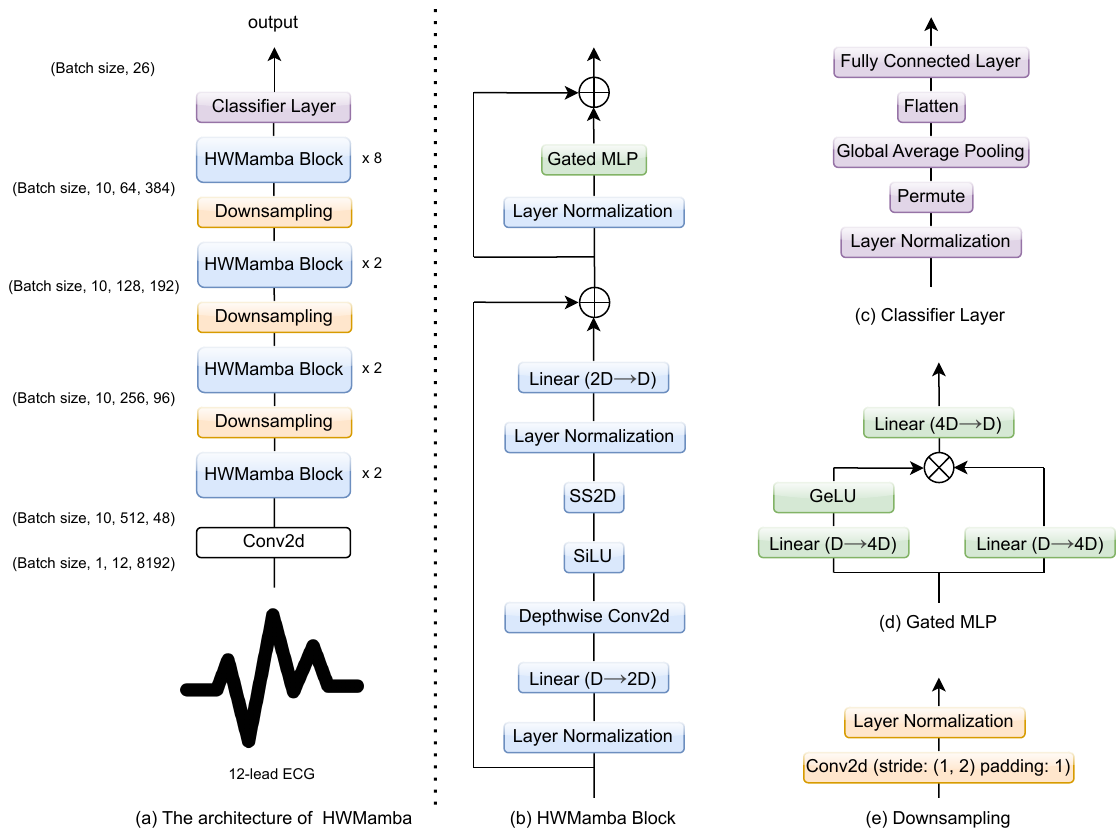}
\caption{HWMamba architecture and submodules. (a) Full model framework; (b) HWMamba block; (c) Classifier layer; (d) Gated MLP; and (e) Downsampling module.} 
\label{arc_HWMamba}
\end{figure}

\subsection*{HexagonalWarriorMamba}
The overall architecture of HWMamba is shown in Figure~\ref{arc_HWMamba}(a). Its hierarchical design is similar to that of the Swin Transformer \cite{liu2021swin} and VMamba~\cite{liu2024vmamba}, processing the input 12-lead ECG at multiple resolutions. This hierarchy is achieved through progressive downsampling, which transitions the representation from high resolution with shallow feature depth to lower resolution with deeper feature depth. 
The Swin Transformer and VMamba operate on square image inputs such as 224×224. In contrast, the proposed model processes the 12-lead ECG as a rectangular input, characterized by a short spatial dimension of 12 leads and a long temporal dimension of 8192 samples. This difference motivates a modification to the initial convolutional layer of HWMamba. The first Conv2d layer utilizes a rectangular kernel of size (3, 16) with a stride of (1, 16) based on insights from 2D-CNN and 2DRU+LC to better accommodate the asymmetric structure of the ECG signals. 

The model then applies multiple HWMamba blocks and downsampling layers. The number of blocks at each stage is empirically determined as [2, 2, 2, 8], with corresponding feature dimensions of [48, 96, 192, 384], as shown in Figure~\ref{arc_HWMamba}(a). A final classifier layer completes the architecture as detailed in Figure~\ref{arc_HWMamba}(c). This layer incorporates layer normalization to stabilize feature distributions, followed by global average pooling and flattening to aggregate spatial information before a fully connected layer produces the final multi-label classification outputs.

\begin{figure}[t]
\centering
\includegraphics[width=\textwidth]{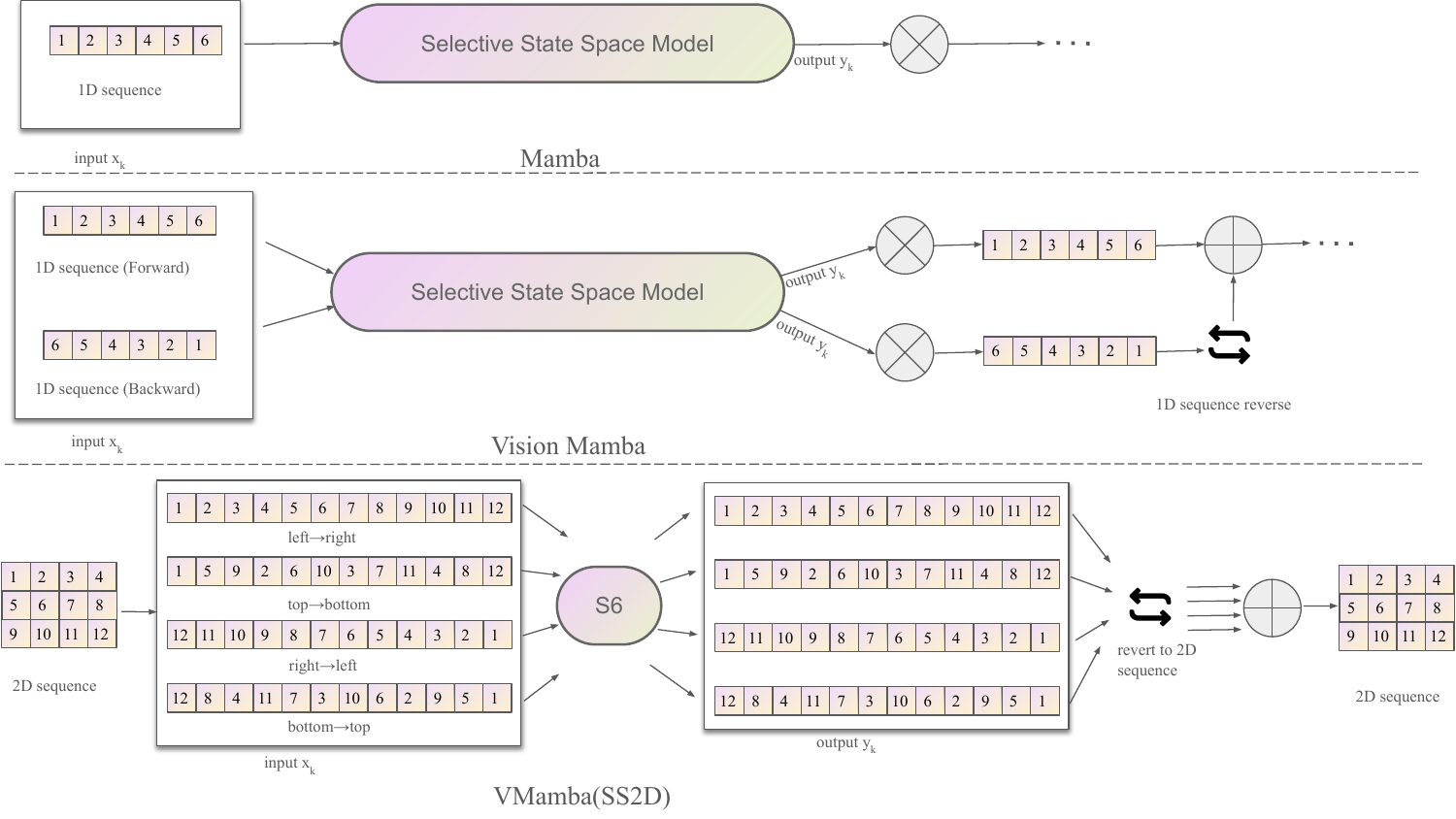}
\caption{Comparison of data sequencing strategies for Selective State Space Models(S6) in Mamba, Vision Mamba and VMamba architectures} 
\label{SSM_driving}
\end{figure}
In Figure~\ref{arc_HWMamba}(b), the SS2D module and the Depthwise Conv2d~\cite{chollet2017xception} are positioned between two linear layers. The SS2D module is a core component of VMamba. As illustrated in Figure~\ref{SSM_driving}, neither Vision Mamba nor VMamba fundamentally alters the underlying Mamba mechanism for processing sequence data; rather, they differ in how they serialize spatial data. 
The standard Mamba operation on an input sequence $\mathbf{X}$ is defined as follows, where $\mathbf{Y}$ represents the output sequence:
\begin{equation}
    \mathbf{Y} = \text{S6}(\mathbf{X})
\end{equation}
Vision Mamba extends this by flattening the feature map and augmenting it with a flipped sequence to scan forward and backward, as expressed by:
\begin{align}
    \mathbf{Y}_{\text{forward}} &= \text{S6}(\mathbf{X}_{\text{forward}}) \\
    \mathbf{Y}_{\text{backward}} &= \text{S6}(\mathbf{X}_{\text{backward}})
\end{align}

In contrast, VMamba transforms the 2D spatial structure into 1D sequences by scanning in four distinct directions (cross-scan). This process is formulated as:
\begin{align}
    \mathbf{Y}_{\text{left}\to\text{right}} &= \text{S6}(\mathbf{X}_{\text{left}\to\text{right}}) \\
    \mathbf{Y}_{\text{top}\to\text{bottom}} &= \text{S6}(\mathbf{X}_{\text{top}\to\text{bottom}}) \\
    \mathbf{Y}_{\text{right}\to\text{left}} &= \text{S6}(\mathbf{X}_{\text{right}\to\text{left}}) \\
    \mathbf{Y}_{\text{bottom}\to\text{top}} &= \text{S6}(\mathbf{X}_{\text{bottom}\to\text{top}})
\end{align}
An SS2D approach based on VMamba is adopted to model spatial dependencies, as the proposed method represents the 12-lead ECG as an image to generate a 2D feature map.
Depthwise Conv2d is employed to reduce computational cost by decomposing standard convolution into two efficient steps. A depthwise convolution first applies one spatial filter per channel independently before a pointwise $1 \times 1$ convolution mixes information across channels. This significantly reduces the number of parameters and computational complexity compared to standard convolutions.

The two linear layers surrounding the SS2D and depthwise convolution form an expansion-contraction MLP block similar to the FFN in Transformers and the MLP block in Vision Mamba/VMamba. This block expands the feature dimension before projecting it back to the original size. As illustrated in Figure~\ref{arc_HWMamba}(b), the channel dimension is expanded by a factor of 2 before reduction. In contrast, the Gated MLP~\cite{liu2021pay, de2024griffin} shown in Figure~\ref{arc_HWMamba}(d) expands the feature dimension by a factor of 4 prior to projection. 
The gating mechanism introduces a multiplicative interaction between features and represents the key reason for using a Gated MLP instead of a standard MLP. It enables adaptive feature selection and contributes to more stable training while preserving informative representations.

Figure~\ref{arc_HWMamba}(e) illustrates the downsampling module, which consists of a Conv2d layer followed by Layer Normalization. The Conv2d uses a $3 \times 3$ kernel. The stride is empirically set to $(1, 2)$ to account for the feature map shown in Figure~\ref{arc_HWMamba}(a). A padding of 1 is applied following the standard downsampling configuration utilized in VMamba. The output is subsequently normalized to the range $[0, 1]$ for evaluation using threshold-dependent metrics.

\section*{Dataset Description}
\label{sec:Dataset}
The experimental data used in this study are sourced from the PhysioNet/CinC Challenge 2021~\cite{physionet_cinc_2021_data, reyna2021will} and consist of 88,253 multi-lead ECGs from seven international institutions. These recordings cover 26 heart conditions (25 abnormalities plus sinus rhythm) and provide multi-label annotations for each. The signals utilize diverse sampling frequencies including 500 Hz (n=87,663) alongside 1000 Hz (n=516) and 257 Hz (n=74). Table \ref{dataset_resulting_list} details the specific subset utilized in this study.



\begin{table}
\centering
\caption{Summary of the ECG datasets sourced from multiple institutions via the PhysioNet/CinC Challenge 2021 utilized in this study.}
\label{dataset_resulting_list}
\begin{tabular}{lcccc}
\toprule
\textbf{Source Database} & \textbf{Recordings ($N$)} & \textbf{Duration (s)} & \textbf{Frequency (Hz)} \\ \midrule
CPSC / CPSC-Extra & 10,330 & 6--144 & 500 \\
St Petersburg INCART & 74 & 1,800 & 257 \\
PTB / PTB-XL & 22,353 & 10--120 & 500 / 1000 \\
Georgia (G12EC) & 10,344 & 5--10 & 500 \\
Ningbo First Hospital & 34,905 & 10 & 500 \\
Chapman / Shaoxing & 10,247 & 10 & 500 \\ \midrule
\textbf{Total Aggregate} & \textbf{88,253} & \textbf{---} & \textbf{---} \\ \bottomrule
\end{tabular}
\end{table}

\section*{Experiments}
\label{sec:ExperimentalSetup}

\subsection*{Experimental Setup}
Both the training and evaluation procedures were implemented using the PyTorch deep learning framework.
All experiments were conducted on a GeForce RTX 3090 Ti GPU with 24 GB of dedicated memory, running Ubuntu 22.04 with CUDA Toolkit 11.8 (Driver 535).

The Adam \cite{kinga2015method} optimizer was used with a base learning rate of $0.001$ and a batch size of $20$. The optimizer parameters were set to $\beta_1 = 0.9$ alongside $\beta_2 = 0.98$ and $\epsilon = 1 \times 10^{-9}$. 
A cosine annealing scheduler \cite{loshchilov2017sgdr} with a warmup phase of 5 epochs was employed to manage the learning rate and align with standard practices in Mamba \cite{gu2024mamba, zhu2024vision, liu2024vmamba} and Transformer \cite{liu2021swin, dosovitskiy2020image} architectures.
This warmup phase linearly increased the learning rate from an initial $1 \times 10^{-5}$ to the base $0.001$. This was followed by cosine decay over 13 epochs, reaching a minimum learning rate of $1 \times 10^{-6}$. The choice of optimizer and scheduler configuration was determined empirically.

The dataset was split into training and testing subsets using stratified proportional allocation and random shuffling. This technique was essential due to the 26 imbalanced classes. The overall data split designated 80\% for training and the remaining 20\% for testing. A 5-fold cross-validation strategy was adopted in place of a separate validation set to provide a more robust evaluation of model performance while maximizing the use of available data. This approach is particularly beneficial for imbalanced datasets. The same training and testing splits were consistently used across experiments to enable direct comparison with SOTA methods. Each fold consisted of 70,602 training records and 17,651 testing records.

A systematic threshold search strategy was applied during the validation phase to evaluate the threshold-dependent performance metrics. The model outputs a probability vector $\hat{y} \in [0, 1]^{C}$ where $C=26$ denotes the number of cardiac conditions. A global decision threshold $\tau$ is used to convert these continuous probabilities into binary predictions. Following standard practice in multi-label classification, a grid search over $\tau \in \{0, 0.02, \dots, 1\}$ was performed. The binary prediction for the $i$-th class, denoted as $\hat{y}_{i,\text{binary}}$, for a given threshold $\tau$, is defined as:

\begin{equation}
\hat{y}_{i, \text{binary}} = 
\begin{cases} 
1 & \text{if } \hat{y}_i > \tau \\
0 & \text{otherwise}
\end{cases}
\end{equation}


\subsection*{Evaluation Metrics}

Six evaluation metrics were employed to compare the proposed HWMamba with SOTA models: Subset Accuracy, Challenge Score, Hamming Loss, Macro F1-score, Weighted F1-score, and Macro AUROC. Among these, AUROC is the only threshold-independent metric, while the remaining five are threshold-dependent.

Individual metrics such as Sensitivity, Specificity, and Precision are not considered primary indicators in this study. These metrics can attain extreme values, for example $1.0$, under trivial threshold settings (e.g., $\tau = 0$ or $\tau = 1$), and may therefore not provide a balanced evaluation. Macro and Weighted F1-scores are adopted instead because they require a trade-off between precision and recall.

Furthermore, AUROC is selected over AUPRC to provide a comprehensive assessment of diagnostic performance. While AUPRC focuses exclusively on the positive class (ignoring true negatives), AUROC incorporates the false positive rate, enabling evaluation of the model's ability to distinguish between positive classes and the negative background. Since the precision-recall trade-off is already reflected by the F1-scores, AUROC serves as a complementary metric for assessing overall ranking performance across both sensitivity and specificity.

\subsubsection*{Subset Accuracy}
Accuracy is commonly used to evaluate the performance of single-label classification tasks. However, in contrast to single-label settings, ECG diagnosis often involves multiple diagnostic labels ~\cite{li2025electrocardiogram}, making it a multi-label classification problem. In this context, subset accuracy provides a more appropriate evaluation metric, as defined below:
\begin{equation}
    Acc_{\text{subset}} = \frac{1}{N} \sum_{i=1}^{N} \mathbb{I}(Y_i = Z_i)
    \label{subset_acc}
\end{equation}
where $Acc_{\text{subset}}$ denotes the subset accuracy metric, $N$ is the total number of samples, and $i$ is the sample index. $Y_i$ represents the true set of labels for sample $i$, while $Z_i$ denotes the predicted set of labels for sample $i$. The indicator function $\mathbb{I}(\cdot)$ equals 1 when $Y_i = Z_i$ (i.e., true match), and 0 otherwise.

\subsubsection*{Challenge Score}

The Challenge Score is a specialized evaluation metric introduced for the PhysioNet/Computing in Cardiology (CinC) 2021 Challenge. It extends the scoring framework used in the 2020 Challenge \cite{alday2020classification} by incorporating a broader range of diagnostic classes and clinically informed weighting. The metric is computed using a predefined weight matrix (e.g., \texttt{weights.csv}), which encodes the clinical relationships between different cardiac conditions.

Unlike standard accuracy, the Challenge Score accounts for the clinical significance of predictions. It assigns partial credit to misclassifications when the predicted condition is clinically similar to the true diagnosis, while applying stronger penalties to errors that may have greater clinical impact. This weighting strategy enables a more clinically meaningful evaluation by reflecting the relative importance of different types of diagnostic errors.
Ultimately, incorporating the Challenge Score metric can bridge deep learning algorithms with clinical realities.

\subsubsection*{Hamming Loss}
The Hamming Loss (Hamming) is an instance-based metric for evaluating multi-label classification models, which measures the fraction of individual labels that are incorrectly predicted. It is computed by comparing the predicted label set with the ground truth set across all samples, counting the total number of label errors and normalizing this count by the total number of labels over all instances. Label-level errors occur when a label is incorrectly predicted as present or absent. The formulation is given below:
\begin{equation} \label{eq:hamming-loss}
    \text{Hamming} = \frac{1}{N \cdot L} \sum_{i=1}^{N} \sum_{j=1}^{L} \mathbb{I}(\hat{y}_{i,j} \neq y_{i,j})
\end{equation}
$N$ is the total number of samples.
$L$ is the total number of labels.
$y_{i,j} \in \{0, 1\}$ is the true label for sample $i$ and label $j$.
$\hat{y}_{i,j} \in \{0, 1\}$ is the predicted label for sample $i$ and label $j$.
$\mathbb{I}(\cdot)$ is the Indicator Function, which returns $1$ if the condition is true (i.e., a misclassification) and $0$ otherwise.

\subsubsection*{Macro F1-score and Weighted F1-score}

The F1-score for an individual label $i$ is the harmonic mean of its Precision ($\text{P}_i$) and Recall ($\text{R}_i$):
\begin{equation}
    \text{F1}_i = \frac{2 \cdot \text{P}_i \cdot \text{R}_i}{\text{P}_i + \text{R}_i}
\end{equation} 

The Macro F1-score ($\text{F1}_{\text{Macro}}$) is calculated as the unweighted arithmetic mean of the individual F1-scores ($\text{F1}_i$) across all $N$ labels, treating each label equally:
\begin{equation} \label{eq:macro-f1}
    \text{F1}_{\text{Macro}} = \frac{1}{N} \sum_{i=1}^{N} \text{F1}_i
\end{equation}


The Weighted F1-score ($\text{F1}_{\text{Weighted}}$) averages the individual F1-scores ($\text{F1}_i$), where each is weighted by the support ($S_i$), which is the number of true instances for label $i$.
\begin{equation} \label{eq:weighted-f1-short}
    \text{F1}_{\text{Weighted}} = \sum_{i=1}^{N} \left( \text{F1}_i \cdot \frac{S_i}{M} \right)
\end{equation}
where $S_i$ is the support (number of true instances) for label $i$, and $M$ is the total number of samples, $M = \sum_{i=1}^{N} S_i$.

\subsubsection*{AUROC}
Macro AUROC is a threshold-independent evaluation metric that assesses a classifier’s ability to distinguish between positive and negative classes. It is computed by averaging the area under the receiver operating characteristic curve across all classes, assigning equal importance to each class regardless of class frequency. This ensures that the model’s overall discriminative capability is evaluated in a balanced manner, even in the presence of class imbalance.

\begin{table}[]
\centering
\caption{Comparison of different HWMamba configurations for multi-label ECG classification. The analysis highlights the impact of classifier architecture (Standard vs. Gated MLP), stride patterns, normalization techniques, and state
dimensions. Best results are \textbf{bolded}, and second-best results are \underline{underlined}.}
\begin{tabular}{@{}lcccccc@{}}
\toprule
\textbf{Configuration}           & \textbf{Subset Accuracy(\%)}        & \textbf{Challenge score} & \textbf{Hamming} $\downarrow$& \textbf{$\text{F1}_{\text{Macro}}$}  & \textbf{$\text{F1}_{\text{Weighted}}$}     & \textbf{AUROC}    \\ 
\midrule
HWMamba (MLP) & \textbf{60.60} & 0.7106 & \textbf{0.0240} & 0.5951 & 0.7603 & 0.9686 \\
HWMamba (Stride: 2,2) & 58.08 & 0.7079 & 0.0266 & 0.5979 & 0.7593 & 0.9672 \\
HWMamba (Stride: 1,3) & 57.37 & 0.6937 & 0.0262 & 0.5682 & 0.7348 & 0.9626 \\
HWMamba (w/ Norm) & 59.06 & 0.7138 & \underline{0.0247} & 0.5982 & 0.7640 & 0.9655 \\
HWMamba (State Dim: 1) & 59.00 & \underline{0.7215} & 0.0252 & \underline{0.5999} & \underline{0.7698} & \textbf{0.9703} \\ \midrule
\textbf{HWMamba (Proposed)} & \underline{59.51} & \textbf{0.7294} & 0.0250 & \textbf{0.6156} & \textbf{0.7730} & \underline{0.9698} \\ 
\bottomrule
\end{tabular}
\label{table_HWMamba_DifferentConfiguration}
\end{table}

\section*{Results and Discussion}
\label{sec:Results}
Table \ref{table_HWMamba_DifferentConfiguration} summarizes the performance of various HWMamba configurations, providing insights into the optimal architectural and preprocessing choices for 12-lead ECG analysis. Unlike the standard VMamba architecture, which is typically designed for 2D image inputs, the experimental results indicate that task-specific adaptations are required for multi-label ECG classification.

The model achieves peak performance with a downsampling stride of (1, 2) and a state dimension of 16. The (1, 2) stride effectively reduces the temporal dimension while preserving critical inter-lead spatial information, outperforming both symmetric (2, 2) and alternative asymmetric (1, 3) strides. Furthermore, utilizing the Gated MLP proved essential for generalization. While a standard MLP yielded higher subset accuracy, the Gated MLP configuration (the final HWMamba) achieved the highest Challenge Score (0.7294) and F1 Weighted (0.7730).

A notable observation relates to the role of input normalization. Although normalization is widely adopted in existing literature as a standard preprocessing step, its application was observed to be suboptimal for this specific task. As observed in the "HWMamba (w/ Norm)" configuration, applying Z-score normalization results in a decrease across most evaluation metrics, with the Challenge Score reduced to 0.7138 and the Macro F1-score to 0.5982 compared to the unnormalized baseline. This suggests that amplitude-related characteristics and the original signal distribution contain discriminative information that is important for the HWMamba model, and may be altered by standard normalization techniques.

\begin{table}[]
\centering
\caption{Comparison of HWMamba (proposed) with SOTA Methods on the PhysioNet/CinC Challenge 2021 Dataset. Best results are \textbf{bolded}, and second-best results are \underline{underlined}.}
\begin{tabular}{@{}lcccccc@{}}
\toprule
\textbf{Method}           & \textbf{Subset Accuracy(\%)}        & \textbf{Challenge score} & \textbf{Hamming} $\downarrow$& \textbf{$\text{F1}_{\text{Macro}}$}  & \textbf{$\text{F1}_{\text{Weighted}}$}     & \textbf{AUROC}    \\
\midrule
ISIBrno & 50.19 & \underline{0.7187} & 0.0365 & 0.5281 & 0.7282 & 0.9048 \\
ISIBrnoWa & 50.42 & 0.6873 & 0.0329 & 0.5119 & 0.7120 & 0.8831 \\
2D-CNN & \underline{57.70} & 0.6726 & 0.0262 & 0.5202 & 0.7235 & 0.9363 \\
2DRU+LC & 56.67 & 0.4659 & \underline{0.0252} & 0.4720 & 0.6860 & \textbf{0.9729} \\ \midrule
ECG-Mamba & 56.10 & 0.7010 & 0.0283 & 0.5731 & \underline{0.7488} & 0.9651 \\
1D-Mamba & 53.15 & 0.6996 & 0.0313 & \underline{0.5887} & 0.7463 & 0.9691 \\
\textbf{HWMamba (Proposed)} & \textbf{59.51} & \textbf{0.7294} & \textbf{0.0250} & \textbf{0.6156} & \textbf{0.7730} & \underline{0.9698} \\  
\bottomrule
\end{tabular}
\label{table_HWMamba_2021}
\end{table}

Table \ref{table_HWMamba_2021} presents a comprehensive quantitative comparison between the proposed HWMamba and current SOTA methods on the PhysioNet/CinC Challenge 2021 dataset. The proposed model was evaluated against three distinct categories of baselines: (1) leading methods from the original challenge, including the winner, ISIBrno, and its updated variant ISIBrnoWa; (2) deep learning approaches based on images, such as 2D CNN and 2DRU+LC, which treat 12-lead ECGs as images with a single channel; and (3) recent Mamba architectures, ECG-Mamba and 1D-Mamba.

To adhere to the principle of no data leakage, the decision thresholds for all threshold-dependent metrics were optimized exclusively on the training set. As shown in Table \ref{table_HWMamba_2021}, HWMamba demonstrates improved performance across threshold-dependent metrics. In particular, the model achieves a Challenge Score of 0.7294, exceeding that of the previous challenge winner ISIBrno (0.7187).
HWMamba also shows consistent performance across complementary metrics, achieving the highest Subset Accuracy (59.51\%), Macro F1-score (0.6156), and Weighted F1-score (0.7730), while maintaining the lowest Hamming Loss (0.0250). These results suggest that the model provides a balanced performance profile across multiple evaluation dimensions.

\begin{figure}[]
\centering
\includegraphics[width=\textwidth]{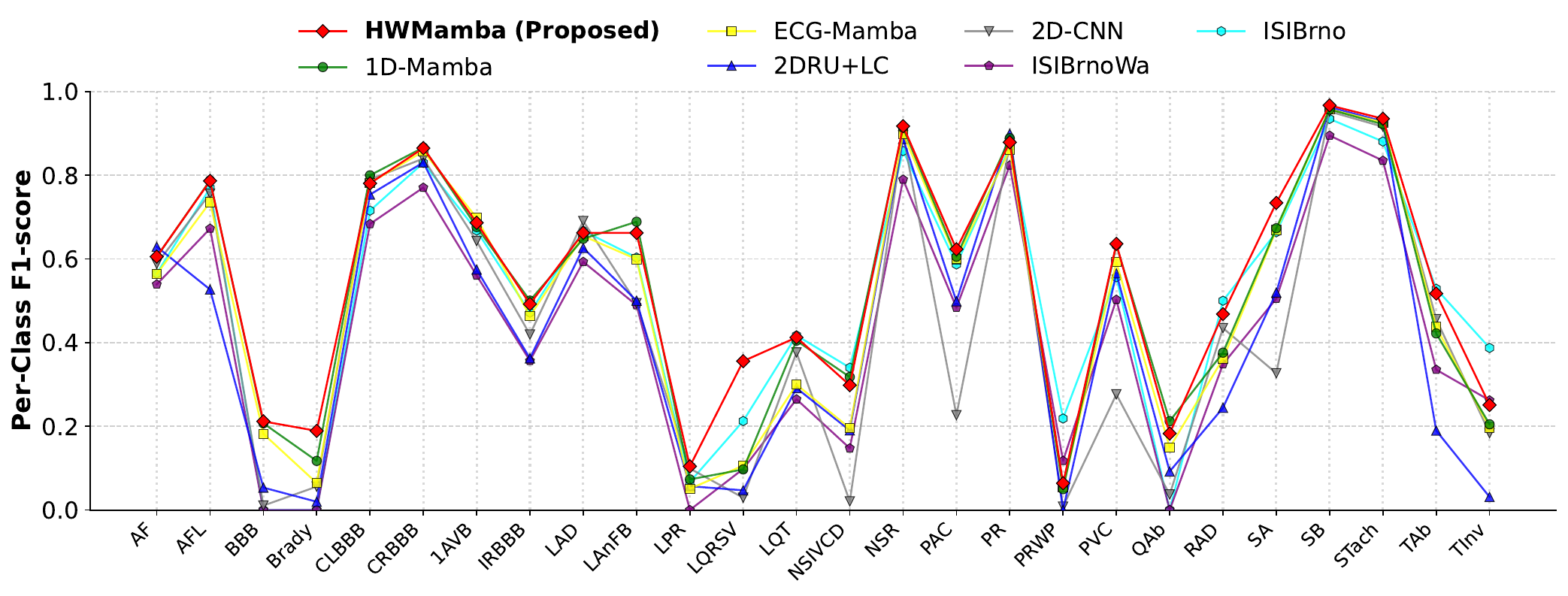}
\caption{Per-Class F1-Score Comparison of HWMamba and SOTA Methods} 
\label{Per_class}
\end{figure}
Figure \ref{Per_class} illustrates the per-class F1-score comparison between HWMamba and SOTA models. The F1-score is chosen because Subset Accuracy, Challenge Score, and Hamming Loss are evaluated at the instance level and cannot assess individual classes. Threshold-independent metrics like AUROC are also excluded to maintain focus on threshold-dependent evaluation. (Abbreviations for diagnosis labels can be found in the Supplementary Information.)
HWMamba consistently outperforms baselines across most cardiac abnormality classes. It achieves F1-scores approaching or exceeding 0.8 in categories like NSR, SB, STach, and CRBBB. Furthermore, HWMamba demonstrates notable resilience in challenging classes such as LPR and LQRSV, where other models struggle. This consistent performance highlights the robustness of the proposed architecture for multi-label ECG classification.

In terms of threshold-independent metrics, the U-Net-adapted method 2DRU+LC remains the SOTA for AUROC (0.9729). However, HWMamba remains highly competitive with an AUROC of 0.9698, trailing by a narrow margin. While 2DRU+LC achieves a marginally higher AUROC, HWMamba significantly outperforms it on all threshold-dependent metrics (Accuracy, F1, and Challenge Score), demonstrating superior capability in making definitive diagnostic predictions.

\begin{figure}[]
    \centering
    \subfloat[Subset Accuracy Comparison: Training vs. Testing Thresholds]{
        \label{BarChart_subset_accuracy} 
        \includegraphics[width=0.48\textwidth]{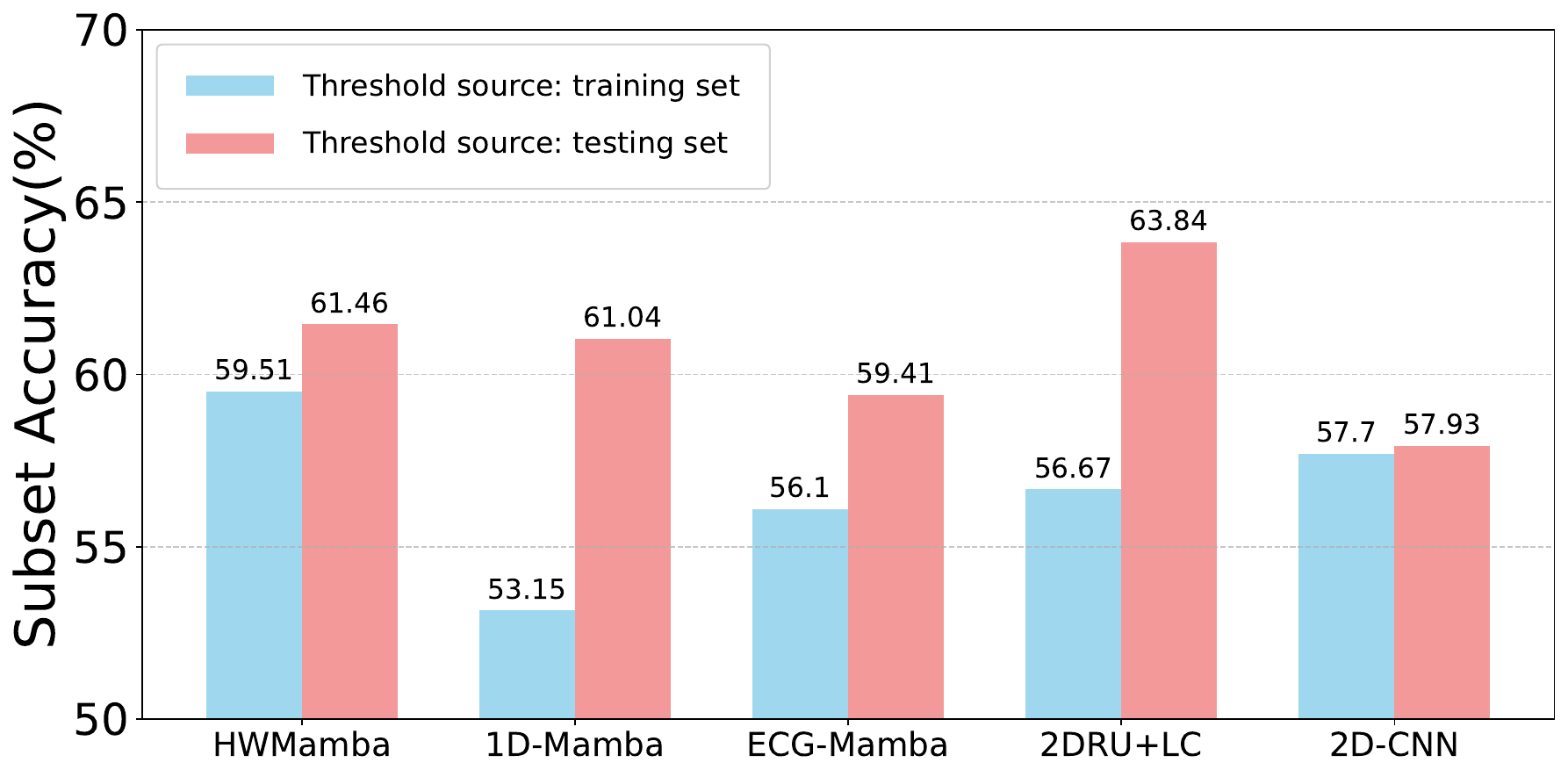}
    }
    \hfill 
    \subfloat[Challenge Score Comparison: Training vs. Testing Thresholds]{
        \label{BarChart_challenge_score} 
        \includegraphics[width=0.48\textwidth]{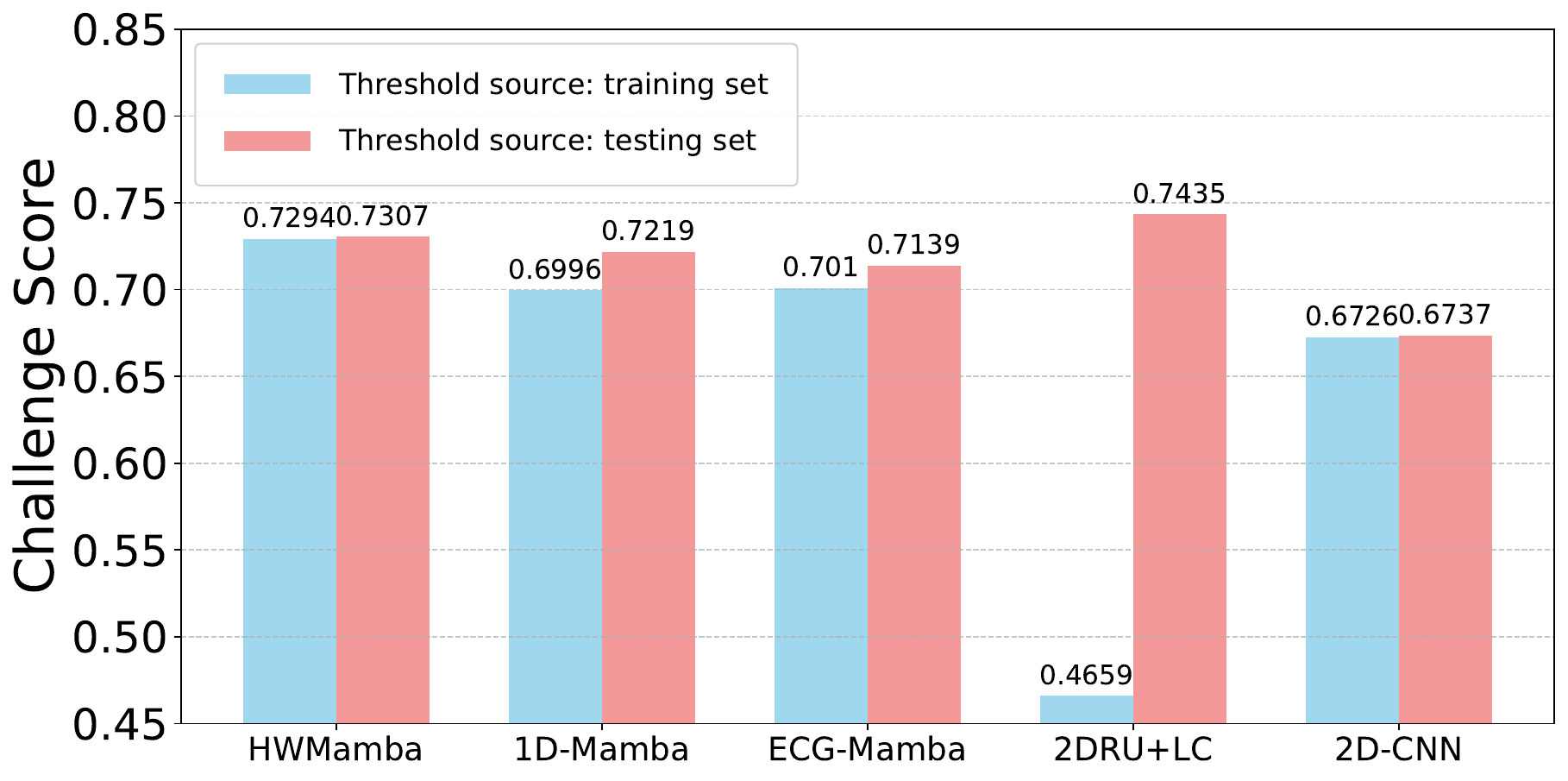} 
    }
    \vfill
    \subfloat[Weighted F1-score Comparison: Training vs. Testing Thresholds]{
        \label{BarChart_weighted_F1-score} 
        \includegraphics[width=0.48\textwidth]{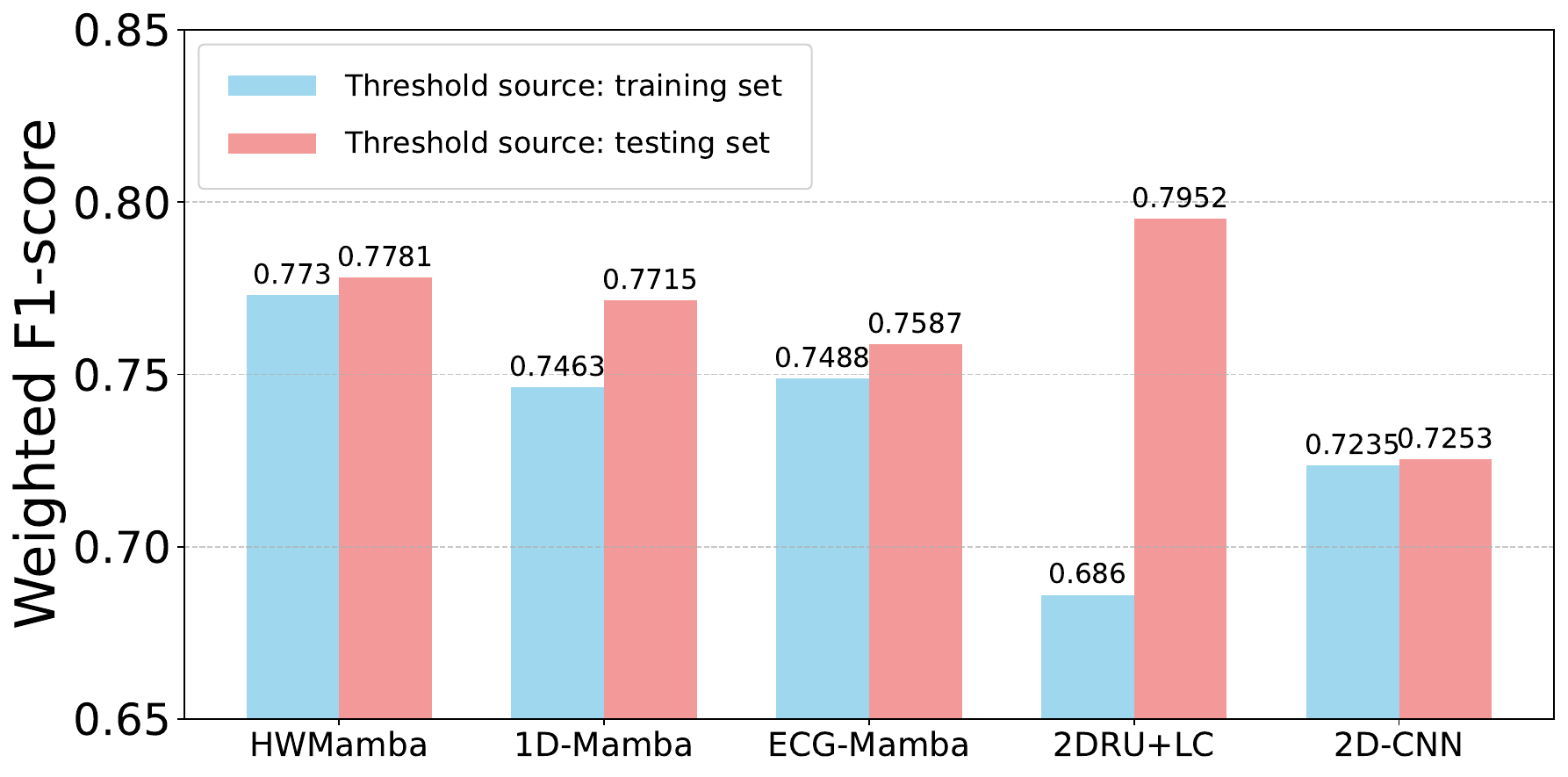} 
    }
    \hfill
    \subfloat[Macro F1-score Comparison: Training vs. Testing Thresholds]{
        \label{BarChart_macro_F1-score} 
        \includegraphics[width=0.48\textwidth]{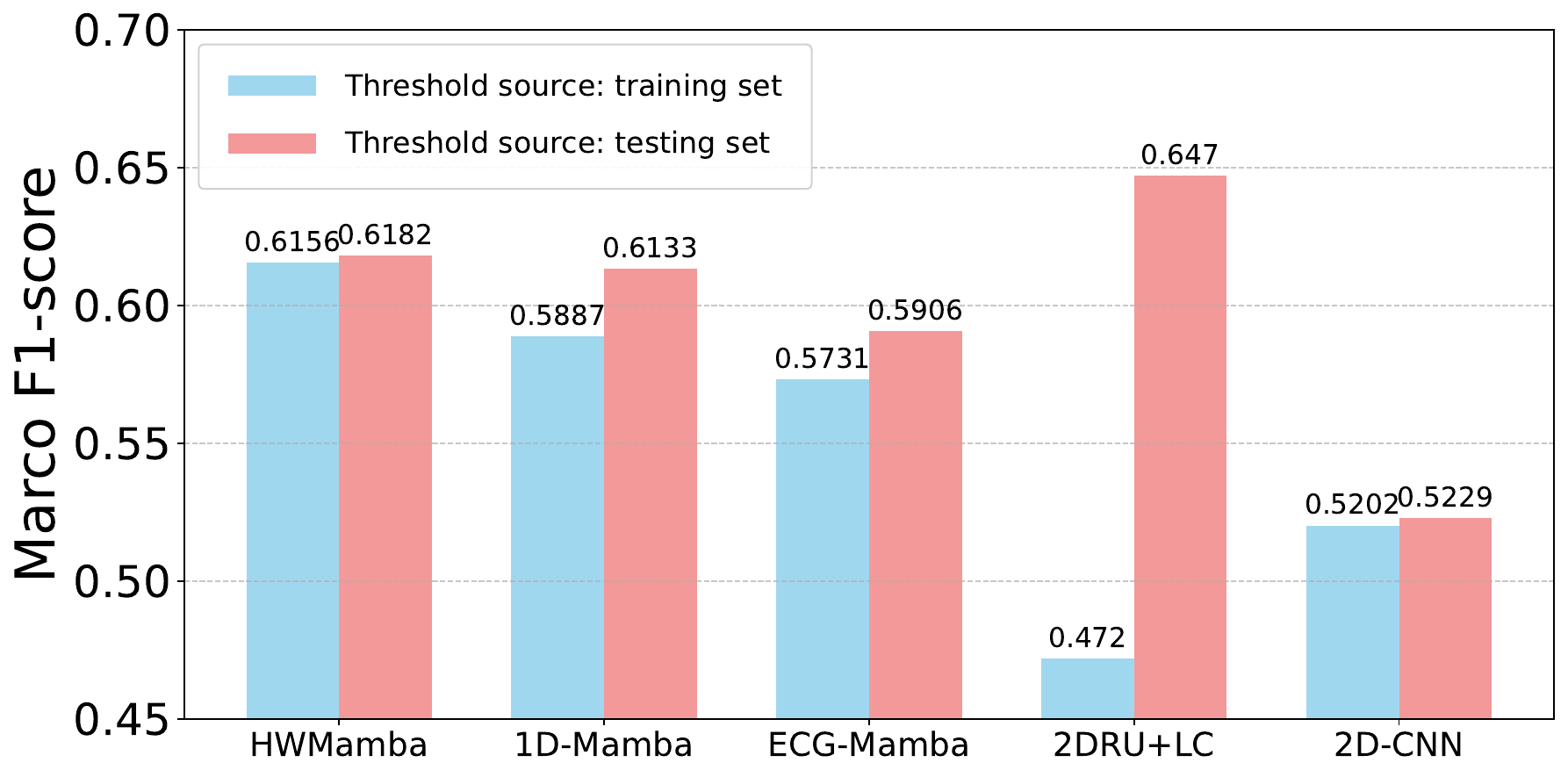} 
    }
    \vfill
    \subfloat[Hamming Loss Comparison: Training vs. Testing Thresholds(lower is better)]{
        \label{BarChart_hamming_loss} 
        \includegraphics[width=0.60\textwidth]{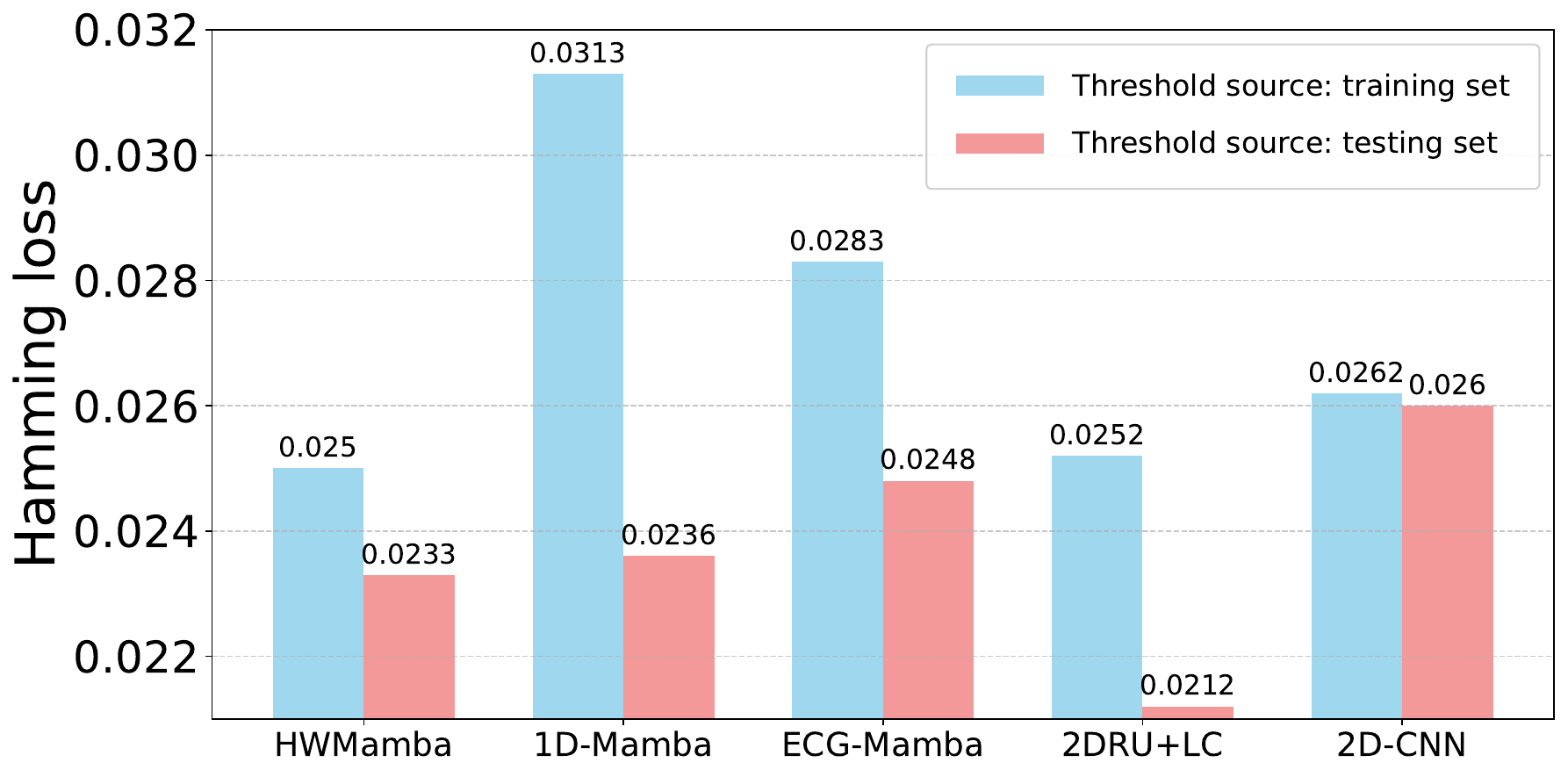} 
    }
    \caption{Impact of Threshold Source on Various Performance Metrics Across Five Models}
    \label{BarChart_5_methods} 
\end{figure}

To examine the distribution characteristics of the HWMamba across the training and testing sets, Figure \ref{BarChart_5_methods} presents a comparative analysis of five models across five threshold-dependent performance metrics: Figure \ref{BarChart_subset_accuracy} Subset Accuracy, Figure \ref{BarChart_challenge_score} Challenge Score, Figure \ref{BarChart_weighted_F1-score} Weighted F1-score, Figure \ref{BarChart_macro_F1-score} Macro F1-score, and Figure \ref{BarChart_hamming_loss} Hamming Loss (where lower values indicate better performance). 
The ISIBrno model and its upgraded variant are excluded from this analysis.
This is because Team ISIBrno uses a custom sparsity loss that constrains the outputs to binary values (0 or 1), resulting in an optimal decision threshold that remains fixed near 1. Consequently, it is not possible to meaningfully analyze threshold-dependent distributional differences between the training and testing sets for this model, unlike those trained using Binary Cross-Entropy.

The bar charts illustrate the impact of the data source used to determine the optimal decision thresholds. The blue bars represent performance when thresholds are calibrated on the training set (realistic scenario), while red bars indicate thresholds optimized on the testing set, representing a theoretical upper bound on performance. The gap between the training and testing thresholds reflects the models' robustness to variations in data distribution between the subsets. A larger disparity indicates decreased robustness and a higher sensitivity to domain shift between training and testing data.

In terms of generalization performance, the 2D-CNN model exhibits the smallest gap between threshold sources across most metrics. Although it does not achieve the highest overall scores, its performance remains stable, indicating strong generalization capabilities and effective handling of distributional consistency between datasets.

In contrast, the 2DRU+LC model demonstrates the largest performance gap, particularly evident in the Challenge Score (Figure \ref{BarChart_challenge_score}) and Hamming Loss (Figure \ref{BarChart_hamming_loss}). 
While performance appears strong using thresholds optimized on test data, the substantially lower results using thresholds derived from training data indicate poor generalization likely caused by sensitivity to distributional mismatches.

The proposed HWMamba method achieves a favorable balance between performance and robustness. 
It consistently delivers competitive results across all evaluated metrics while maintaining a relatively small gap between threshold settings derived from training and testing.
This indicates that the model not only achieves strong predictive performance but also demonstrates improved resilience to distributional variation between datasets.

\section*{Conclusions}
This study introduced the HWMamba framework to redefine the automated diagnosis of cardiac abnormalities by processing 12-lead ECG signals as 2D representations with a single channel. Integrating a hierarchical architecture with the SS2D mechanism allows HWMamba to successfully overcome the limitations of conventional 1D time series models in capturing dependencies over long ranges and complex spatial relationships within ECG data.
The comprehensive evaluation on the large-scale PhysioNet/CinC Challenge 2021 dataset demonstrates that HWMamba establishes a benchmark for multi-label ECG classification. The model achieved SOTA performance across five critical threshold-dependent metrics, including Subset Accuracy, Challenge Score, and Weighted F1-score while maintaining near-SOTA performance in AUROC. This balanced performance profile highlights the model’s ability to simultaneously optimize multiple evaluation criteria, indicating both high predictive accuracy and robustness in multi-label cardiac diagnosis.

Further analysis reveals that standard input normalization leads to a measurable degradation in performance, suggesting that absolute ECG amplitude carries diagnostically relevant information that is effectively preserved and exploited by HWMamba. In addition, the relatively small performance gap between training-derived and testing-derived decision thresholds indicates strong generalization capability, which is critical for reliable deployment in clinical settings.
%
In contrast, the 2DRU+LC approach based on the U-Net architecture exhibits limited generalization when thresholds are derived from the training set, despite achieving strong performance under optimally tuned thresholds.
This observation underscores the sensitivity of certain architectures to distributional shifts and highlights an important direction for future research. In particular, combining the structural advantages of U-Net with the sequence modeling capabilities of Mamba-based architectures may provide a promising pathway to improve both threshold robustness and overall classification performance.


\section*{Data availability}
The datasets analyzed during the current study were sourced from the Official PhysioNet/CinC Challenge 2021. The data is publicly available in the PhysioNet repository and can be accessed at the following stable URL: \url{https://physionet.org/content/challenge-2021/1.0.3/}.

\bibliography{sample}



\section*{Author contributions statement}
H.J. conceived the original idea, designed and performed the simulations, and wrote the main manuscript. V.S. and M.S.M. developed and finalized the mathematical modeling for the State Space Model. D.R., W.P., and S.W. interpreted the 12-lead ECGs for heart disease detection. J.L. prepared the figures and graphical representations. H.M. and J.Y. reviewed and edited the manuscript, significantly improving its technical clarity. All authors reviewed and approved the final manuscript.

\section*{Funding}
This study was supported by the National Research Foundation of Korea (NRF), funded by the Korean government (Ministry of Science and ICT) (RS-2025-00519038).

\section*{Additional information}
The authors declare no competing interests.

\section*{Supplementary information}
This section provides the detailed mathematical derivation and proof for the discretization process of the State Space Model (SSM) utilized in the HexagonalWarriorMamba framework. Additionally, it defines the nomenclature and variables essential to this process as summarized in Table~\ref{tab:ssm_nomenclature} and details the abbreviations for each diagnosis label in Table~\ref{abbre_per_class}.

\subsection*{Derivation of SSM Discretization}

The Mamba architecture transforms the continuous-time State Space Model (SSM) into a discrete-time formulation suitable for deep learning sequences.

Consider the continuous-time system:
\begin{align}
    h'(t) &= \mathbf{A}h(t) + \mathbf{B}x(t) \label{eq:state_equation} \\
    y(t) &= \mathbf{C}h(t)
\end{align}
where $\mathbf{A}$ and $\mathbf{B}$ are constant matrices.

\subsection*{Step 1: General Solution via Laplace Transform}
Since the state equation~\eqref{eq:state_equation} is a first-order differential equation, the Laplace transform is applied. For a function $f(t)$, the transform is defined as:
\begin{equation}
    \mathcal{L}\{ f(t) \} = H(s) = \int_{0}^{\infty} f(t) e^{-st} \, dt 
    \label{Laplace_definition}
\end{equation}

The transform of the derivative $h'(t)$ is derived using integration by parts:
\begin{equation}
    \mathcal{L}\{ h'(t) \} = \int_{0}^{\infty} h'(t) e^{-st} \, dt
\end{equation}
Let $u = e^{-st}$ and $dv = h'(t)dt$. Then $du = -s e^{-st}dt$ and $v = h(t)$. Applying the formula $\int u \, dv = uv - \int v \, du$:
\begin{equation}
    \int_{0}^{\infty} h'(t) e^{-st} \, dt = \left[ h(t) e^{-st} \right]_{0}^{\infty} + s \int_{0}^{\infty} h(t) e^{-st} \, dt
    \label{Laplace_outcome}
\end{equation}
Assuming system stability (where $\lim_{t \to \infty} h(t)e^{-st} = 0$), the boundary term simplifies to $-h(0)$. Substituting the definition of $H(s)$:
\begin{equation}
    \mathcal{L}\{ h'(t) \} = s H(s) - h(0)
\end{equation}

Next, the linearity of the Laplace transform is applied to the right-hand side of Eq.~\eqref{eq:state_equation}:
\begin{equation}
    \mathcal{L}\{ \mathbf{A} h(t) + \mathbf{B} x(t) \} = \mathbf{A} H(s) + \mathbf{B} X(s)
\end{equation}

Equating the transformed sides:
\begin{equation} 
    s H(s) - h(0) = \mathbf{A} H(s) + \mathbf{B} X(s) 
\end{equation}

Rearranging to solve for $H(s)$ (using the identity matrix $\mathbf{I}$):
\begin{equation} 
    (s\mathbf{I} - \mathbf{A}) H(s) = h(0) + \mathbf{B} X(s)
\end{equation}
\begin{equation}
    H(s) = (s\mathbf{I} - \mathbf{A})^{-1} h(0) + (s\mathbf{I} - \mathbf{A})^{-1} \mathbf{B}X(s)
    \label{Laplace_outcome4}
\end{equation}

The time-domain solution $h(t)$ is recovered via the inverse Laplace transform:
\begin{enumerate}
    \item \textbf{First term:} Recognized as the matrix exponential: $\mathcal{L}^{-1} \{ (s\mathbf{I} - \mathbf{A})^{-1} \} = e^{\mathbf{A}t}$.
    \item \textbf{Second term:} Handled via the Convolution Theorem, $\mathcal{L}^{-1} \{ F(s) G(s) \} = (f * g)(t)$.
\end{enumerate}

Thus, the general solution is:
\begin{equation}
    h(t) = e^{\mathbf{A}t} h(0) + \int_{0}^{t} e^{\mathbf{A}(t - \tau)} \mathbf{B} x(\tau) \, d\tau
\end{equation}

Generalizing the initial time from $0$ to an arbitrary $t_0$ yields the Variation of Constants formula:
\begin{equation}
    h(t) = e^{\mathbf{A}(t - t_0)} h(t_0) + \int_{t_0}^{t} e^{\mathbf{A}(t - \tau)} \mathbf{B} x(\tau) \, d\tau
    \label{Master_formula}
\end{equation}

\subsection*{Step 2: Discrete-Time Formulation}
To discretize the system, let the time steps be defined by a fixed step size $\Delta$, such that:
\begin{equation}
    t_0 = t_{k-1}, \quad t = t_k, \quad \text{and} \quad \Delta = t_k - t_{k-1}
\end{equation}
Substituting these into Eq.~\eqref{Master_formula}:
\begin{equation}
    h(t_k) = e^{\mathbf{A}\Delta} h(t_{k-1}) + \int_{t_{k-1}}^{t_k} e^{\mathbf{A}(t_k - \tau)} \mathbf{B} x(\tau) \, d\tau
\end{equation}

\subsection*{Step 3: Zero-Order Hold Discretization}
The analytical evaluation of the integral requires knowledge of $x(\tau)$ between discrete steps. The Zero-Order Hold assumption is applied, where the input is assumed constant over the interval $[t_{k-1}, t_k]$:
\begin{equation}
    x(\tau) \approx x_k \quad \text{for} \quad \tau \in [t_{k-1}, t_k]
\end{equation}
Since $x_k$ and $\mathbf{B}$ are constant with respect to $\tau$, they are extracted from the integral:
\begin{equation}
    h_k = e^{\mathbf{A}\Delta} h_{k-1} + \left( \int_{t_{k-1}}^{t_k} e^{\mathbf{A}(t_k - \tau)} \, d\tau \right) \mathbf{B} x_k
\end{equation}

\subsection*{Step 4: Evaluation of the Integral} 
The integral term is solved using the change of variables $u = t_k - \tau$, implying $du = -d\tau$. The limits change from $[t_{k-1}, t_k]$ to $[\Delta, 0]$:
\begin{equation}
    \int_{t_{k-1}}^{t_k} e^{\mathbf{A}(t_k - \tau)} \, d\tau = \int_{\Delta}^{0} e^{\mathbf{A}u} (-du) = \int_{0}^{\Delta} e^{\mathbf{A}u} \, du
\end{equation}
Assuming $\mathbf{A}$ is invertible, the integral of the matrix exponential is:
\begin{equation}
    \left[ \mathbf{A}^{-1} e^{\mathbf{A}u} \right]_{0}^{\Delta} = \mathbf{A}^{-1} (e^{\mathbf{A}\Delta} - \mathbf{I})
\end{equation}

\subsection*{Final Discrete Recurrence} 
Substituting the integral result back into the state equation yields the discrete recurrence rule:
\begin{equation}
    h_k = \underbrace{e^{\mathbf{A}\Delta}}_{\overline{\mathbf{A}}} h_{k-1} + \underbrace{\mathbf{A}^{-1}(e^{\mathbf{A}\Delta} - \mathbf{I}) \mathbf{B}}_{\overline{\mathbf{B}}} x_k
    \label{assembly_fun}
\end{equation}

The discretized system parameters $\overline{\mathbf{A}}$ and $\overline{\mathbf{B}}$ are formally defined as:
\begin{align}
    \overline{\mathbf{A}} &= \exp(\mathbf{A}\Delta) \\
    \overline{\mathbf{B}} &= \mathbf{A}^{-1}(\exp(\mathbf{A}\Delta) - \mathbf{I}) \cdot \mathbf{B} 
    \label{discretized_B}
\end{align}
resulting in the discrete linear recurrence:
\begin{equation}
    h_k = \overline{\mathbf{A}} h_{k-1} + \overline{\mathbf{B}} x_k
\end{equation}

\begin{figure}
\centering
\includegraphics[width=0.5\textwidth]{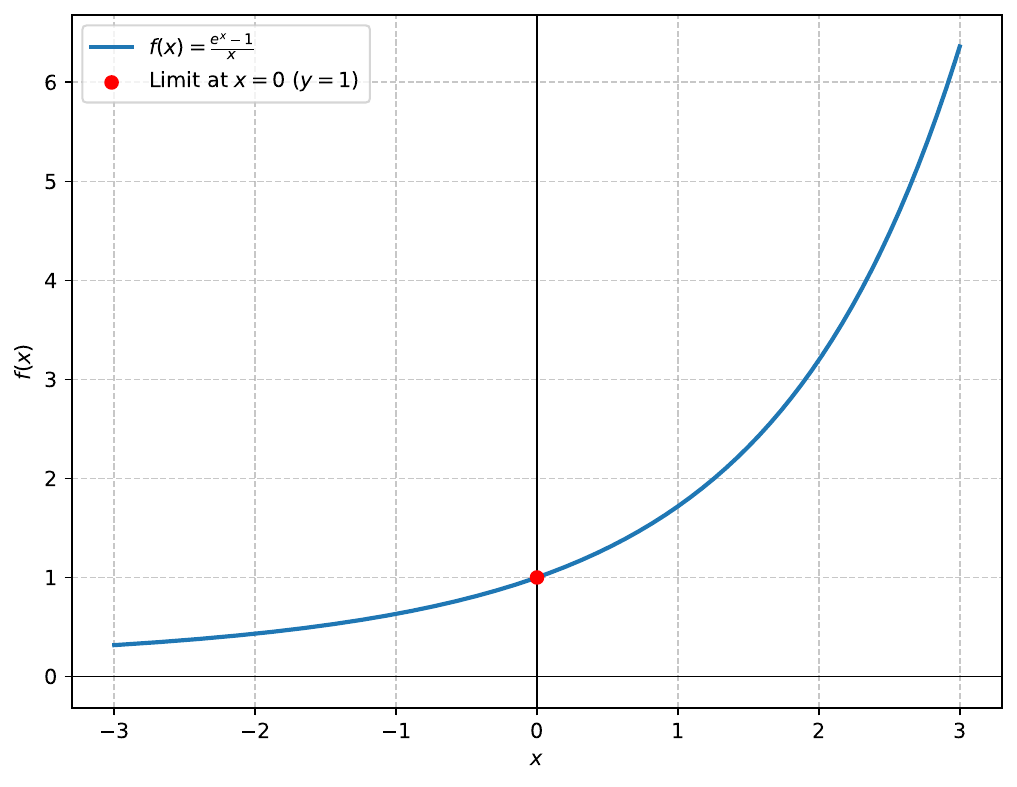}
\caption{Graph of the Function $f(x) = \frac{e^x - 1}{x}$} 
\label{Bernoulli's_rule}
\end{figure}

Equation~\eqref{discretized_B} implies an inversion of $\mathbf{A}$, which can be conceptually related to the scalar function:
\begin{equation}
    f(x) = \frac{e^x - 1}{x}
    \label{discretized_B2}
\end{equation}

As illustrated in Figure~\ref{Bernoulli's_rule}, this function $f(x)$ remains strictly positive for all real $x$. However, regarding the matrix term $\mathbf{A}^{-1}$, if $\mathbf{A}$ possesses eigenvalues close to zero (singular or ill-conditioned), direct computation of the inverse leads to numerical instability or undefined values (NaN).

To resolve this, the expression is reformulated to avoid direct inversion of small values. The term $(\bm{\Delta} \mathbf{A})^{-1} (\exp(\bm{\Delta} \mathbf{A}) - \mathbf{I})$ corresponds to the scalar function $f(x)$ where $x = \bm{\Delta} \mathbf{A}$. Since $\lim_{x \to 0} \frac{e^x - 1}{x} = 1$, this form remains numerically stable even as $\mathbf{A} \to 0$.

By multiplying the numerator and denominator by $\bm{\Delta}$, the discretization is rewritten in its robust form:
\begin{align}
    \overline{\mathbf{A}} &= \exp(\bm{\Delta} \mathbf{A}) \\
    \overline{\mathbf{B}} &= (\bm{\Delta} \mathbf{A})^{-1}(\exp(\bm{\Delta} \mathbf{A}) - \mathbf{I}) \cdot (\bm{\Delta} \mathbf{B})
\end{align}

\begin{table}[h]
    \centering
    \renewcommand{\arraystretch}{1.2} 
    \caption{Nomenclature and Variable Definitions for SSM Discretization}
    \label{tab:ssm_nomenclature}
    \begin{tabular}{l p{10cm}}
        \hline
        \textbf{Symbol} & \textbf{Description} \\
        \hline
        \multicolumn{2}{l}{\textit{Continuous-Time System Parameters}} \\
        $\mathbf{A}$ & Continuous-time state transition matrix \\
        $\mathbf{B}$ & Continuous-time input projection matrix \\
        $\mathbf{C}$ & Output projection matrix \\
        $\mathbf{I}$ & Identity matrix \\
        
        \hline
        \multicolumn{2}{l}{\textit{Time-Domain Variables}} \\
        $t$ & Continuous time variable \\
        $h(t)$ & Continuous-time hidden state vector \\
        $h'(t)$ & Time derivative of the hidden state \\
        $x(t)$ & Continuous-time input signal \\
        $y(t)$ & Output signal \\
        $t_0$ & Arbitrary initial time \\
        $\tau$ & Integration variable (dummy time variable) \\
        
        \hline
        \multicolumn{2}{l}{\textit{Laplace Domain}} \\
        $s$ & Complex frequency variable in the Laplace domain \\
        $\mathcal{L}\{ \cdot \}$ & Laplace transform operator \\
        $H(s)$ & Laplace transform of the hidden state $h(t)$ \\
        $X(s)$ & Laplace transform of the input $x(t)$ \\
        
        \hline
        \multicolumn{2}{l}{\textit{Discretization Parameters}} \\
        $\Delta$ & Discrete time step size (sampling interval) \\
        $k$ & Discrete time step index \\
        $t_k$ & Discrete time point at step $k$ \\
        $h_k$ & Discrete hidden state vector at step $k$ \\
        $x_k$ & Discrete input vector at step $k$ (Zero-Order Hold) \\
        $\overline{\mathbf{A}}$ & Discretized state transition matrix ($\exp(\mathbf{A}\Delta)$) \\
        $\overline{\mathbf{B}}$ & Discretized input matrix \\
        $f(x)$ & Scalar function $\frac{e^x - 1}{x}$ used for numerical stability \\
        \hline
    \end{tabular}
\end{table}

\pagebreak

\subsection*{Class-Wise Performance Analysis}

\begin{table}[htbp]
\centering
\caption{PhysioNet/Computing in Cardiology Challenge 2021 Scored Diagnoses}
\label{tab:cinc2021_diagnoses}
\begin{tabular}{ll}
\hline
\textbf{Abbreviation} & \textbf{Diagnosis Label} \\
\hline
AF & Atrial Fibrillation \\
AFL & Atrial Flutter \\
BBB & Bundle Branch Block \\
Brady & Bradycardia \\
CLBBB & Complete Left Bundle Branch Block \\
CRBBB & Complete Right Bundle Branch Block \\
1AVB & 1st Degree Atrioventricular Block \\
IRBBB & Incomplete Right Bundle Branch Block \\
LAD & Left Axis Deviation \\
LAnFB & Left Anterior Fascicular Block \\
LPR & Prolonged PR Interval \\
LQRSV & Low QRS Voltages \\
LQT & Prolonged QT Interval \\
NSIVCD & Nonspecific Intraventricular Conduction Disorder \\
NSR & Sinus Rhythm \\
PAC & Premature Atrial Contraction \\
PR & Pacing Rhythm \\
PRWP & Poor R Wave Progression \\
PVC & Premature Ventricular Contractions \\
QAb & Q Wave Abnormal \\
RAD & Right Axis Deviation \\
SA & Sinus Arrhythmia \\
SB & Sinus Bradycardia \\
STach & Sinus Tachycardia \\
TAb & T Wave Abnormal \\
TInv & T Wave Inversion \\
\hline
\end{tabular}
\label{abbre_per_class}
\end{table}

\end{document}